\documentclass[runningheads]{llncs}
\usepackage{graphicx}
\graphicspath{ {./figures/} }
\usepackage{multirow}

\usepackage{bbm}

\usepackage{amsfonts,amssymb}

\usepackage[normalem]{ulem}
\useunder{\uline}{\ul}{}

\begin{document}
%
\title{Aspect-Sentiment-Multiple-Opinion Triplet Extraction}
\titlerunning{Aspect-Sentiment-Multiple-Opinion Triplet Extraction}
%
\author{
	Fang Wang\inst{2}\textsuperscript{($\star$)} \and
	Yuncong Li\inst{1}\textsuperscript{($\star$)} \and
	Sheng-hua Zhong\inst{2}\textsuperscript{($\dagger$)}
	\and
	Cunxiang Yin\inst{1}\textsuperscript{}
	\and
	Yancheng He\inst{1}\textsuperscript{}
}
%
\authorrunning{F. Wang et al.}
%
\institute{Tencent Inc., Shenzhen, China\\
	\email{\{liyuncong,jasonyin,collinhe\}@tencent.com} \and
	College of Computer Science and Software Engineering, Shenzhen University, Shenzhen, China\\
	\email{wangfang20161@email.szu.edu.cn}, \email{csshzhong@szu.edu.cn}}

\renewcommand{\thefootnote}{\fnsymbol{footnote}}
\footnotetext[1]{Equal contribution.}
\footnotetext[4]{Corresponding author.}

\renewcommand{\thefootnote}{\arabic{footnote}}

\maketitle              
\begin{abstract}
Aspect Sentiment Triplet Extraction (ASTE) aims to extract aspect term (aspect), sentiment and opinion term (opinion) triplets from sentences and can tell a complete story, i.e., the discussed aspect, the sentiment toward the aspect, and the cause of the sentiment. ASTE is a charming task, however, one triplet extracted by ASTE only includes one opinion of the aspect, but an aspect in a sentence may have multiple corresponding opinions and one opinion only provides part of the reason why the aspect has this sentiment, as a consequence, some triplets extracted by ASTE are hard to understand, and provide erroneous information for downstream tasks. In this paper, we introduce a new task, named Aspect Sentiment Multiple Opinions Triplet Extraction (ASMOTE). ASMOTE aims to extract aspect, sentiment and multiple opinions triplets. Specifically, one triplet extracted by ASMOTE contains all opinions about the aspect and can tell the exact reason that the aspect has the sentiment. We propose an Aspect-Guided Framework (AGF) to address this task. AGF first extracts aspects, then predicts their opinions and sentiments. Moreover, with the help of the proposed Sequence Labeling Attention(SLA), AGF improves the performance of the sentiment classification using the extracted opinions. Experimental results on multiple datasets demonstrate the effectiveness of our approach 
\footnote{Data and code can be found at https://github.com/l294265421/ASMOTE}.

\end{abstract}
\section{Introduction}
\begin{figure}
	\centering
	\includegraphics[scale=0.15]{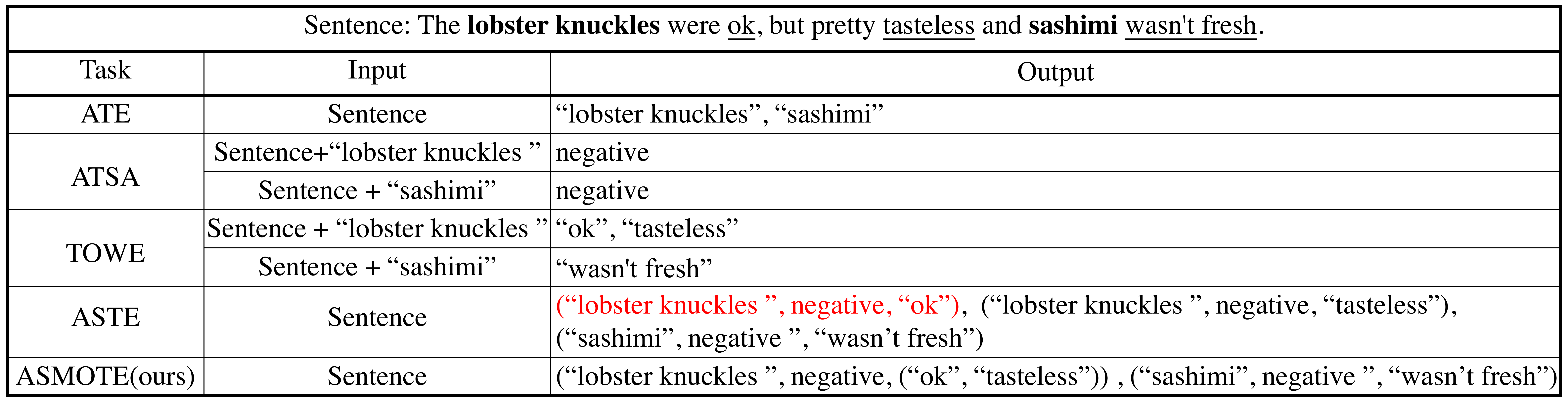}
	\caption{\label{fig:examples}An example of our ASMOTE and several ABSA subtasks. In the sentence, the bold words are aspects and the underlined words are opinions. The red triplet extracted by ASTE is confusing, because the sentiment of ``ok'' is neutral rather than negative.}
\end{figure}

Sentiment analysis~\cite{pang2008opinion,liu2012sentiment} is an important task in natural language understanding and receives much attention in recent years. Aspect-based sentiment analysis (ABSA) \cite{pontiki-etal-2014-semeval,pontiki-etal-2015-semeval,pontiki-etal-2016-semeval} is a branch of sentiment analysis. ABSA includes several subtasks, such as Aspect Term Extraction (ATE), Aspect Term Sentiment Analysis (ATSA) and Target-oriented Opinion Words Extraction (TOWE) \cite{fan2019target}. \textbf{Aspect terms} (or simply \textbf{aspects}) are the linguistic expressions used in sentences to refer to the reviewed entities. \textbf{Opinion terms}  (or simply \textbf{opinions}) are the expressions that carry subjective attitudes in sentences. Given a sentence, ATE extracts the aspects in the sentence. Given a sentence and an aspect in the sentence, ATSA and TOWE predict the corresponding sentiment and opinions respectively. For example, given the sentence in Fig.~\ref{fig:examples}, ATE extracts ``lobster knuckles'' and ``sashimi''. ATSA predicts the negative sentiments toward ``lobster knuckles'' and ``sashimi''. TOWE extracts ``ok'' and ``tasteless'' for ``lobster knuckles'' and ``wasn't fresh'' for ``sashimi''. 

The individual subtasks mentioned above or a combination of two subtasks can only answer one question or two questions, but can not tell a complete story, i.e. the discussed aspect, the sentiment toward the aspect, and the cause of the sentiment. To address this limitation, Peng et al.~\cite{Peng2020KnowingWH} introduced the Aspect Sentiment Triplet Extraction (ASTE) task. A triplet extracted from a sentence by ASTE includes an aspect, the sentiment that the sentence expresses toward the aspect, and one opinion about the aspect in the sentence. Given the sentence in Fig.~\ref{fig:examples}, ASTE extracts three triplets, (``lobster knuckles'', negative, ``ok''), (``lobster knuckles'', negative, ``tasteless'') and  (``sashimi'', negative, ``wasn't fresh'').

However, one triplet extracted by ASTE only includes one opinion of the aspect, but an aspect in a sentence may have multiple corresponding opinions. One opinion of the aspect with multiple opinions only provides part of the reason why the aspect has the sentiment, resulting in some triplets extracted by ASTE are hard to understand and provide erroneous information for downstream tasks. For example, when seeing the triplet,  (``lobster knuckles'', negative, ``ok''), extracted by ASTE, we will be confused, because the sentiment of ``ok'' is neutral. 

One direct solution to the problem of ASTE is to add a post-processing step after ASTE models, which combines the multiple triplets with the same aspect into one triplet. For example, the post-processing step merges the triplets extracted by ASTE from the sentence in Fig.~\ref{fig:examples} and obtains two triplets: (``lobster knuckles'', negative, (``ok'', ``tasteless'')) and (``sashimi'', negative, (``wasn't fresh'')). The obtained triplets are correct and understandable, since they contains all the opinions in the sentence about the aspect and the opinions in these triplets can tell the exact reason that the aspect has the sentiment. However, this solution of patching is not elegant. And, ASTE models need to  extract  erroneous triplets (e.g. (``lobster knuckles'', negative, ``ok'')), which is unreasonable.

In this paper, we introduce a new task, Aspect Sentiment Multiple Opinions Triplet Extraction (ASMOTE). ASMOTE has the same goal as the combination of ASTE and the post-processing step. That is,  ASMOTE extracts aspect, sentiment and multiple opinions triplets. One triplet extracted by ASMOTE contains all the opinions in the sentence about the aspect. The example illustrated in  Fig.~\ref{fig:examples} shows the inputs and outputs of the tasks mentioned above.

We propose an Aspect-Guided Framework (AGF) for ASMOTE. AGF includes two stages. The first stage extracts aspects, and the second stage predicts the sentiments and opinions of the aspects extracted in the first stage. The ASMOTE triplets can be obtained by merging the results of the two stages. Specifically, given a sentence, the first stage uses a neural sequence labeling model to extract aspects. For each aspect extracted in the first stage, AGF generates aspect-specific representations with the guidance of the aspect. The obtained representations are used to predict the corresponding sentiment and opinions of the aspect. AGF also uses a neural sequence labeling model to extract opinions associated with the aspect. Moreover, it is intuitive that the opinions of an aspect can help models predict the sentiment of the aspect. For example, given the sentence in Fig.~\ref{fig:examples} and the aspect ``sashimi'', if AGF knows that the opinion associated with ``sashimi'' is  the phrase ``wasn't fresh'', AGF will predict the negative sentiment more easily. Based on the intuition, we propose a Sequence Labeling Attention(SLA). Specifically, SLA converts the prediction results of the neural sequence labeling model for opinion extraction into attention weights. The attention weights are used to generate an opinion representation. The opinion representation is used to predict the sentiment of the aspect. SLA is a kind of attention mechanism with supervision and sequence labeling tasks can be seen as attention supervision tasks. 

Our contributions are summarized as follows:
\begin{itemize}
	\item We introduce a new task, Aspect Sentiment Multiple Opinions Triplet Extraction (ASMOTE).
	\item  We propose an Aspect-Guided Framework (AGF) for ASMOTE and a Sequence Labeling Attention (SLA). AGF improves the performance of the sentiment classification using extracted opinions with the help of SLA.
	\item Experimental results on four public datasets demonstrate the effectiveness of AGF and SLA.
\end{itemize}

\section{Related Work}
Aspect-based sentiment analysis (ABSA)~\cite{pontiki-etal-2014-semeval,pontiki-etal-2015-semeval,pontiki-etal-2016-semeval} aims to address various sentiment analysis tasks at a fine-grained level. ABSA includes several subtasks, such as Aspect Term Extraction (ATE), Opinion Term Extraction (OTE) extracting opinions from sentences and Aspect Term Sentiment Analysis (ATSA). Many methods have been proposed for these subtasks, such as~\cite{li2018aspect,xu2018double,wei2020don} for ATE, \cite{wang2016recursive,wang2017coupled,dai2019neural} for OTE and \cite{dong2014adaptive,tang2020dependency,wang2020relational,zhao2020modeling} for ATSA. Since the three subtasks are correlated in pairs, some studies improved the performances of the three subtasks by jointly modelling two or three of them. Wang et al.~\cite{wang2016recursive,wang2017coupled} and Dai
and Song~\cite{dai2019neural} jointly modelled ATE and OTE. Li et al.~\cite{li2019unified} and Phan and Ogunbona~\cite{phan2020modelling} jointly modelled ATE and ATSA. He et al.~\cite{he2019interactive} and Chen and Qian~\cite{chen2020relation} jointly modelled ATE, OTE and ATSA. 

Although extracting aspects and opinions as pairs is significant, the aspects and opinions extracted by the methods mentioned above are not in pairs \cite{fan2019target}. Fan et al.~\cite{fan2019target} put forward a new subtask of ABSA: Target-oriented Opinion Words Extraction (TOWE). TOWE aims to extract the corresponding opinions with respect to the given aspect. A few methods \cite{fan2019target,wu2020latent} have been proposed for TOWE. While TOWE assumes the golden aspect was given, Zhao et al.~\cite{zhao2020spanmlt} and Chen et al.~\cite{chen2020synchronous} explored Aspect-Opinion Pair extraction task, which aims at extracting aspects and opinions in pairs without given golden aspects. 

The above tasks are still not enough to get a complete picture regarding sentiment \cite{Peng2020KnowingWH}. Peng et al.~\cite{Peng2020KnowingWH} proposed a new subtask: Aspect Sentiment Triplet Extraction (ASTE). ASTE extracts aspect, sentiment and opinion triplets and can tell a complete story, i.e. the discussed aspect, the sentiment toward the aspect, and the cause of the sentiment. Some methods~\cite{Peng2020KnowingWH,xu-etal-2020-position,10.1007/978-3-030-60450-9_52,zhang-etal-2020-multi-task,wu-etal-2020-grid} have been proposed for ASTE. However, ASTE has the problem mentioned in Introduction section \footnote{There is a contemporaneous work~\cite{li2021finegrained}, which also proposes a task, Aspect-Sentiment-Opinion Triplet Extraction (ASOTE), to solve the problem of ASTE.}.

\section{Aspect-Guided Framework (AGF)}

\subsection{Framework}
\begin{figure*}
 	\centering
 	\includegraphics[width=0.99\textwidth]{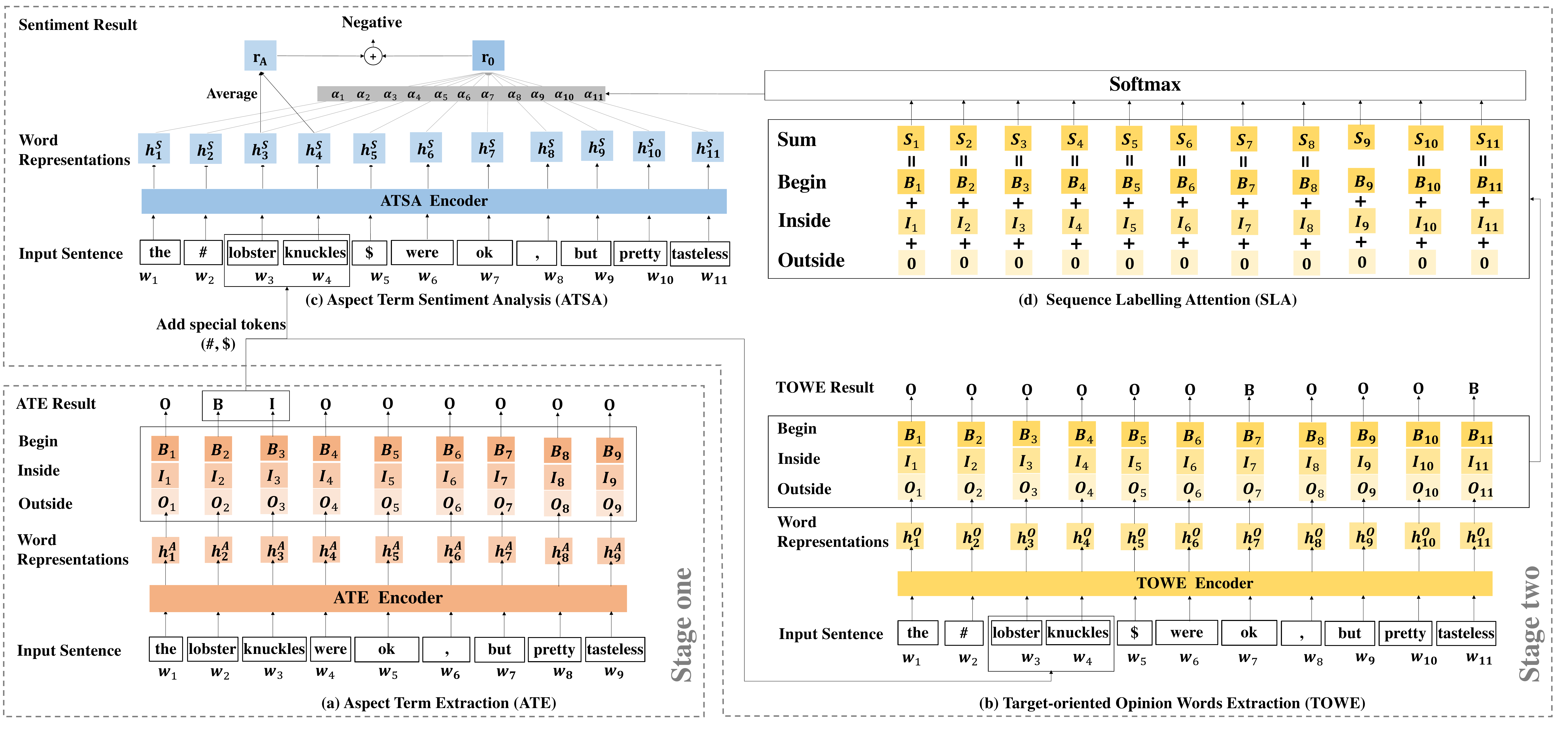}
 	\caption{Our proposed Aspect-Guided Framework (AGF) for ASMOTE.}
 	\label{fig:model}
 \end{figure*}

The overall architecture of our Aspect-Guided Framework (AGF) for ASMOTE is shown in Fig.~\ref{fig:model}. AGF decomposes ASMOTE into three subtasks: Aspect Term Extraction (ATE), Aspect Term Sentiment Analysis (ATSA) and Target-oriented Opinion Words Extraction (TOWE). Given a sentence $S=\{w_1,...,w_i,\\...,w_n\}$, ATE extracts a set of aspects $A=\{a_1,...,a_j,...,a_m\}$. For each aspect extracted by ATE, $a_j$, ATSA predicts its sentiment $s_j \in \{positive, neutral, \\negative\}$, and TOWE extracts its opinions $O=\{o_j^1,...,o_j^k,..., o_j^{l_j}\}$. An aspect may have more than one opinion and $l_j$ is the number of opinions with respect to the $j$-th aspect. ASMOTE obtains the triplets by merging the results of the three subtasks: $T=\{(a_1, s_1, (o_1^1, ..., o_1^{l_1})),...,(a_m, s_m, (o_m^1, ..., o_m^{l_m}))\}$.

 AGF can be divided into two stages. The first stage performs ATE and the second stage performs ATSA and TOWE jointly. Moreover, the aspects extracted in the first stage are used to guide the sentence encoders of ATSA and TOWE to generate aspect-specific sentence representations. 
 
\subsection{Encoders}
Note that AGF is a general framework, we can use any network as the encoder to learn sentence representations for ATE or aspect-specific sentence representations for ATSA and TOWE. In this paper, we implement three different encoders. The first one is the BiLSTM with pre-trained word embeddings, which has been widely used in neural-based models for NLP tasks. The second is BERT~\cite{devlin2019bert}, a pre-trained bidirectional transformer encoder, which has achieved state-of-the-art performances across a variety of NLP tasks. The third is the BiLSTM with BERT, which has been widely used in neural sequence labeling models. The three encoders are written as $Encoder_{BiLSTM\_EMB}$, $Encoder_{BERT}$, and $Encoder_{BiLSTM\_BERT}$, respectively. All the three encoders take a sentence, $S=\{w_1,...,w_i,...,w_n\}$, as input, and output corresponding sentence representations, $H=\{h_1,...,h_i,...,h_n\}$.

\subsection{Stage One: Aspect Term Extraction (ATE)}
We formulate ATE as a sequence labeling problem. Given a sentence $S=\{w_1,...,w_i,...,w_n\}$, an encoder takes the sentence as input and outputs the corresponding sentence representation, $H^A=\{h^A_1,...,h^A_i,...,h^A_n\}$. ATE uses $h^A_i$ to predict the tag $y^A_i \in \{B, I, O\}$ of the word $w_i$. It can be regarded as a three-class classification problem at
each position of the sentence $S$. We use a linear layer and a softmax layer to compute prediction probability $\hat{y}^A_i$:
\begin{equation}
\hat{y}^A_i=softmax(W^A_1h^A_i + b^A_1)
\end{equation}
where $W^A_1$ and $b^A_1$ are learnable parameters.

The cross-entropy loss of ATE task can be defined as follows:
\begin{equation}
L_{ATE}=-\sum_{i=1}^{n}\sum_{t=0}^{2}\mathbb{I}(y^A_i=t)log(\hat{y}^A_{i_t})
\end{equation}
where the tags $\{B, I, O\}$ are correspondingly converted into labels $\{0, 1, 2\}$ and $y^A_i$ denotes the ground truth label. $\mathbb{I}$ is an indicator function.  If $y^A_i==t$, $\mathbb{I}$ = 1, otherwise 0. We minimize the loss $L_{ATE}$ to optimize the ATE model.

\subsection{Stage Two}
Since, in the same sentence, for different aspects, ATSA and TOWE models need to output different  results. Therefore, it is crucial for both ATSA models and TOWE models to learn aspect-specific sentence representations. Moreover, Liang et al.~\cite{liang2019novel} and Xing et al.~\cite{xing2019earlier} showed that utilizing the given aspect to guide the sentence encoding can obtain better aspect-specific representations. Therefore, in AGF, the aspects extracted in the first stage are used to guide the sentence encoders of ATSA and TOWE to generate aspect-specific sentence representations. 
\subsubsection{Target-oriented Opinion Words Extraction (TOWE).}
\label{TOWE}
We also formulate TOWE as a sequence labeling problem. Given a sentence $S=\{w_1,...,w_i,...,w_n\}$ and an aspect in the sentence, we first modify the sentence by inserting the special token \# at the beginning of the aspect and the special token \$ at the end of the aspect. We then get a new sentence  $S_{new}=\{w_1,..., \#, w_{a_s},..., w_{a_e}, \$,...,w_{n+2}\}$, where $\{w_{a_s},..., w_{a_e}\}$ are the corresponding words with respect to the given aspect. An encoder takes the new sentence as input. The special tokens explicitly tell the sentence encoder the corresponding words of the aspect in the sentence. Special tokens were first used by Wu and He~\cite{wu2019enriching} to incorporate target entities information into BERT on the relation classification task. We explore this method in not only BERT-based models but also LSTM-based models. The sentence encoder outputs the aspect-specific sentence representation,  $H^O=\{h^O_1,..., h^O_i,...,h^O_{n+2}\}$. TOWE uses $h^O_i$ to predict the tag $y^O_i \in \{B, I, O\}$ of the word $w_i$ in the new sentence. It can be regarded as a three-class classification problem at
each position of $S_{new}$. We use a linear layer and a softmax layer to compute logit $\hat{l}^O_i$ and its probability $\hat{y}^O_i$:
\begin{equation}
\hat{l}^O_i=W^O_1h^O_i + b^O_1, \hat{y}^O_i=softmax(\hat{l}^O_i)
\end{equation}
where $W^O_1$ and $b^O_1$ are learnable parameters.

The cross-entropy loss of TOWE task can be defined as follows:
\begin{equation}
L_{TOWE}=-\sum_{i=1}^{n}\sum_{t=0}^{2}\mathbb{I}(y^O_i=t)log(\hat{y}^O_{i_t})
\end{equation}
where the tags $\{B, I, O\}$ are correspondingly converted into labels $\{0, 1, 2\}$ and $y^O_i$ denotes the ground truth label.

\subsubsection{Sequence Labeling Attention (SLA).}
SLA converts the prediction results of TOWE into an attention vector. Specifically, the input of SLA is logits $\{\hat{l}^O_1,..., \hat{l}^O_i,...,  \hat{l}^O_{n+2}\}$. SLA then obtains a vector: 
\begin{equation}
\beta=[\beta_1, ..., \beta_i, ..., \beta_{n+2}]
\end{equation}
where $\beta_i$ is computed by summing the predicted logits on TOWE related to labels B and I in $\hat{l}^O_i$. SLA uses the softmax function on $\beta$ to get the attention weight vector:
\begin{equation}
\alpha=[\alpha_1, ..., \alpha_i,  ..., \alpha_{n+2}]
\end{equation}
SLA can take probabilities $\{\hat{y}^O_1,..., \hat{y}^O_i,..., \hat{y}^O_{n+2}\}$ as input. When taking probabilities as input, SLA will behave differently.

\subsubsection{Aspect Term Sentiment Analysis (ATSA).}
We formulate ATSA as a text span-based classification problem.  In this paper, we use the given aspect to guide the sentence encoding in a similar manner as the TOWE task. Given a sentence $S=\{w_1,...,w_i,...,w_n\}$ and an aspect in the sentence, we first modify the sentence by inserting the special token \# at the beginning of the aspect and the special token \$ at the end of the aspect. We then get a new sentence $S_{new}=\{w_1,...,\#, w_{a_s},..., w_{a_e}, \$,...,w_{n+2}\}$, where $\{w_{a_s},..., w_{a_e}\}$ are the corresponding words with respect to the given aspect. An encoder takes the new sentence as input and outputs the corresponding sentence representation, $H^S=\{h^S_1,..., h^S_{a_s},..., h^S_{a_e},...,h^S_{n+2}\}$. We then obtain the aspect representation by averaging the corresponding hidden states:
\begin{equation}
r_A=\frac{1}{(a_e-a_s+1)}\sum_{i=a_e}^{i=a_s}h^S_i
\end{equation}

We use the attention vector $\alpha$ generated by SLA to obtain the opinion representation:
\begin{equation}
r_O=H^S\alpha^T
\end{equation}

AGF concatenates $r_A$ with $r_O$ to get the aspect-specific sentence representation for ATSA:
\begin{equation}
r=[r_A;r_O]
\end{equation}

The aspect-specific representation is then used to predict the sentiment polarity of the aspect. Formally, its sentiment distribution is calculated by:
\begin{equation}
p=softmax(W^S_2(ReLU(W^S_1r+b^S_1))+b^S_2)
\end{equation}
where $W^S_1$, $b^S_1$, $W^S_2$ and $b^S_2$ are parameters.

We use cross entropy as the loss function:
\begin{equation}
L_{ATSA}=-\sum_{t=0}^{2}\mathbb{I}(y^S=t)logp_t
\end{equation}
where $y^S$ denotes the ground truth label and the sentiments $\{positive, neutral, \\negative\}$ are correspondingly converted into labels $\{0, 1, 2\}$. 

\subsubsection{Loss.}
The loss of the second stage is defined as follows:
\begin{equation}
L_{second}=L_{TOWE} + L_{ATSA}
\end{equation}
We minimize the loss $L_{second}$ to optimize the ATSA and TOWE model.

\section{Experiments}
\subsection{Datasets and Metrics}
We construct four datasets (i.e., 14res, 14lap, 15res, 16res) to evaluate the performance of methods on the ASMOTE task. Similar to Peng et al.~\cite{Peng2020KnowingWH} who constructed the Aspect Sentiment Triplet Extraction (ASTE) datasets, we obtain the four ASMOTE datasets by aligning the four Target-oriented Opinion Words Extraction (TOWE) datasets~\cite{fan2019target} and the corresponding SemEval Challenge datasets~\cite{pontiki-etal-2014-semeval,pontiki-etal-2015-semeval,pontiki-etal-2016-semeval}. We do not use the ASTE datasets constructed by previous studies~\cite{Peng2020KnowingWH,xu-etal-2020-position} to build ASMOTE datasets (i.e., combine the multiple triplets with the same aspect in the ASTE datasets into one triplet to get ASMOTE triplets), because these datasets do not include the sentences which only contain aspects without corresponding opinion terms. We think datasets including these sentences can better evaluate the performance of methods, since methods can encounter this kind of sentences in real-world scenarios. Statistics of the ASMOTE datasets are given in Table~\ref{tabel:ASMOTE-data}. Since the number of triplets with conflict sentiment is small, these triplets are removed in our experiments.

\begin{table}
	\caption{\label{tabel:ASMOTE-data} Dataset statistics. The tc indicates triplet with conflict sentiment.}
	\centering
	\begin{tabular}{|l|l|l|l|l|l|l|l|l|l|l|l|l|}
		\hline
		\multicolumn{1}{|c|}{\multirow{2}{*}{Dataset}} & \multicolumn{3}{l|}{14res} & \multicolumn{3}{l|}{14lap} & \multicolumn{3}{l|}{15res} & \multicolumn{3}{l|}{16res} \\ \cline{2-13} 
		\multicolumn{1}{|c|}{}                         & train    & dev    & test   & train    & dev    & test   & train    & dev    & test   & train    & dev    & test   \\ \hline
		\#sentence                                      & 1615     & 404    & 606    & 1183     & 296    & 422    & 666      & 167    & 401    & 987      & 247    & 419    \\ \hline
		\#aspect                                        & 2943     & 751    & 1134   & 1883     & 482    & 656    & 961      & 238    & 542    & 1391     & 352    & 612    \\ \hline
		\#triplet                                       & 2116     & 522    & 864    & 1295     & 315    & 481    & 870      & 206    & 436    & 1206     & 301    & 456    \\ \hline
		\#tc                                            & 74       & 12     & 13     & 31       & 5      & 14     & 7        & 2      & 6      & 17       & 1      & 8      \\ \hline
	\end{tabular}
\end{table}

To evaluate the performance of methods on ASMOTE, we use precision, recall, and F1-score as the metrics. A extracted triplet is regarded as correct only if predicted aspect spans, sentiment, multiple opinions spans and ground truth aspect spans, sentiment, multiple opinions spans are exactly matched. 

\subsection{Our Methods}
\textbf{AGF} uses the encoder $Encoder_{BiLSTM\_EMB}$ for ATE, TOWE and ATSA.

\textbf{AGF-p} is the pipeline version of AGF. AGF-p doesn't contain SLA and performs ATSA and TOWE separately.

\textbf{AGF-t} is a variant of AGF. AGF-t replaces the TOWE model jointly trained in AGF with the TOWE model trained separately. That is, AGF-t only uses the results of ATSA jointly trained in AGF.

\textbf{*$_S$} are variants of AGF*. *$_S$ indicate that SLA takes probabilites rather than logits as input.

\textbf{*$^B$} use encoder $Encoder_{BERT}$ for ATSA, and encoder $Encoder_{BiLSTM\_BERT}$ for both ATE and TOWE. The parameters of BERT are fixed during training.

\textbf{*$^{BF}$} are variants of *$^B$. *$^{BF}$ finetune BERT during training.

\subsection{Implementation Details}
We implement our models in PyTorch \cite{paszke2017automatic}. We use 300-dimensional word vectors pretrained by GloVe \cite{pennington2014glove} to initialize the word embedding vectors. $Encoder_{BERT}$ and $Encoder_{BiLSTM\_BERT}$ use the uncased basic pre-trained BERT. The batch size is set to 32 for all models. All models are optimized by the Adam optimizer \cite{kingma2014adam}. The learning rates are set to 0.001 and 0.00002 for non-BERT models and BERT-based models, respectively. Since the TOWE model is harder to converge than the ATSA model, for AGF, TOWE is trained first then both of TOWE and ATSA are trained together. We apply early stopping in training and the patience is 10. We run all models for 5 times and report the average results on the test datasets.

\begin{table*}
	\caption{Results of the ASMOTE task.}\label{table:asmote}
	\centering
	\begin{tabular}{|c|ccc|ccc|ccc|ccc|}
		\hline
		& \multicolumn{3}{c|}{14res}                          & \multicolumn{3}{c|}{14lap}                          & \multicolumn{3}{c|}{15res}                          & \multicolumn{3}{c|}{16res}                          \\ \hline
		Method                                              & P               & R               & F1              & P               & R               & F1              & P               & R               & F1              & P               & R               & F1              \\ \hline
		MTL & 56.7 & 38.6 & 45.9 & 41.4 & 24.2 & 30.5 & 57.2 & 33.5 & 42.1 & 54.1 & 44.8 & 49 \\
		JET$^t (M=6)$ & 47.1 & 47.8 & 47.4 & 42.4 & 33.8 & 37.6 & 55.7 & 40.7 & 47.0 & 56.2 & 49.6 & 52.7 \\ 
		JET$^o (M=6)$ & 58.6 & 48.6 & 53.2 & 39.9 & 31.2 & 35.0 & 52.7 & 42.6 & 47.1 & \textbf{62.3} & 56.5 & 59.3 \\ 
		GTS-CNN                                             & 59.8          & 51.3           & 55.2          & \textbf{46.3} & 34.7          & 39.7           & 51.6          & \textbf{45.7} & 48.4          & 54.9          & 57.0          & 55.9          \\
		GTS-BiLSTM                                          & 60.4          & 45.8          & 52.1          & 46.0          & 30.3          & 36.5          & \textbf{62.2} & 40.9           & 49.3          & 61.8          & 48.9          & 54.6          \\
		AGF-t (ours)         & \textbf{62.8} & \textbf{56.7} & \textbf{59.6} & 46.1          & \textbf{35.0} & \textbf{39.8}  & 55.9          & 45.3          & \textbf{50.0} & 62.1 & \textbf{58.8} & \textbf{60.4} \\ \hline
		JET$^t_{+bert} (M=6)$ & 50.3 & 53.0 & 51.6 & 44.8 & 35.3 & 39.5 & 55.4 & 44.2 & 49.2 & 51.4 & 56.9 & 54.0 \\
		JET$^o_{+bert} (M=6)$ & 57.0 & 47.6 & 51.9 & 43.0 & 33.5 & 37.7 & \textbf{58.0} & 47.0 & 51.9 & \textbf{66.7} & 54.5 & 60.0 \\
		GTS-BERT                                            & \textbf{63.9} & 61.6          & 62.7          & \textbf{51.7} & 44.6          & 47.9           & 57.9 & 53.3          & 55.5          & 56.4          & 64.5          & 60.2          \\
		AGF-t$^{BF}$ (ours) & 63.5           & \textbf{64.6}  & \textbf{64.0} & 46.7          & \textbf{49.3} & \textbf{48.0} & 56.8          & \textbf{54.6} & \textbf{55.7}  & 57.7 & \textbf{69.5} & \textbf{63.0} \\ \hline
	\end{tabular}
\end{table*}

\subsection{Comparison Methods}
We compare our methods with sevaral methods proposed for the ASTE task, including i) five non-BERT models: MTL~\cite{zhang-etal-2020-multi-task}, JET$^t$~\cite{xu-etal-2020-position}, JET$^o$~\cite{xu-etal-2020-position}, GTS-CNN~\cite{wu-etal-2020-grid}, and GTS-BiLSTM~\cite{wu-etal-2020-grid}, ii) three BERT-based models: JET$^t_{+bert}$~\cite{xu-etal-2020-position}, JET$^o_{+bert}$~\cite{xu-etal-2020-position} and GTS-BERT~\cite{wu-etal-2020-grid}. We add a post-processing step described in Introduction section after these models
to obtain ASMOTE triplets.

\subsection{Results}
Experimental results of our methods and baselines on the ASMOTE task are reported in Table~\ref{table:asmote}. From
Table~\ref{table:asmote} we draw the following conclusions. First, although all the baselines are joint models, which jointly extract ASTE triplets, our pipeline models, AGF-t and AGF-t$^{BF}$, achieve better F1 score than their counterparts, indicating that, to achieve ASMOTE, well-designed models for ASMOTE is more effective than a combination of an ASTE model and the post-processing step. Second, AGF-t$^{BF}$ outperforms AGF-t on all datasets in terms of F1 acore, which shows that BERT can boost the performance of AGF-t.

\begin{table}[]
	\caption{Results of the variants of AGF on ASMOTE, ATSA and TOWE.}\label{table:as}
	\centering
	\begin{tabular}{|c|cccc|cccc|cccc|}
		\hline
		\multirow{2}{*}{Method} & \multicolumn{4}{c|}{ASMOTE (F1)}                                                                     & \multicolumn{4}{c|}{ATSA (accuracy)}                                                                       & \multicolumn{4}{c|}{TOWE (F1)}                                                                             \\ \cline{2-13} 
		& \multicolumn{1}{c|}{14res} & \multicolumn{1}{c|}{14lap} & \multicolumn{1}{c|}{15res} & 16res         & \multicolumn{1}{c|}{14res} & \multicolumn{1}{c|}{14lap} & \multicolumn{1}{c|}{15res} & 16res               & \multicolumn{1}{c|}{14res} & \multicolumn{1}{c|}{14lap} & \multicolumn{1}{c|}{15res} & 16res               \\ \hline
		AGF-p                   & 57.5                       & 39.7                       & 48.0                       & 58.7          & 77.2                       & 69.8                       & 73.9                       & 86.3                & 77.1                       & \textbf{68.0}              & 69.9                       & \textbf{79.6}       \\ \cline{1-1}
		AGF                     & {\ul 59.5}                 & 38.4                       & 48.8                       & {\ul 58.0}    & {\ul \textbf{80.5}}        & {\ul \textbf{72.0}}        & {\ul \textbf{76.7}}        & {\ul \textbf{87.7}} & 76.8                       & 66.2                       & 69.8                       & 77.2                \\ \cline{1-1}
		AGF$_S$                 & 57.9                       & {\ul 38.9}                 & {\ul 50.0}                 & 57.4          & 78.3                       & 68.9                       & 75.7                       & 86.3                & {\ul \textbf{77.4}}        & {\ul 67.2}                 & {\ul \textbf{71.1}}        & {\ul 78.2}          \\ \cline{1-1}
		AGF-t                   & \textbf{59.6}              & \textbf{39.8}              & \textbf{50.0}              & \textbf{60.4} & —                          & —                          & —                          & —                   & —                          & —                          & —                          & —                   \\ \hline
		AGF-p$^B$               & 55.4                       & 40.6                       & 50.2                       & 56.2          & 73.8                       & 64.3                       & 71.0                       & 82.4                & 79.7                       & 70.4                       & \textbf{76.5}              & \textbf{81.6}       \\ \cline{1-1}
		AGF$^B$                 & {\ul 59.3}                 & 45.1                       & {\ul 53.9}                 & {\ul 58.5}    & {\ul \textbf{80.8}}        & {\ul \textbf{76.1}}        & {\ul \textbf{81.7}}        & {\ul \textbf{88.8}} & 78.2                       & 70.3                       & 75.1                       & {\ul 80.5}          \\ \cline{1-1}
		AGF$_S^B$               & 58.4                       & {\ul \textbf{45.8}}        & 53.5                       & 56.8          & 79.7                       & 74.0                       & 77.9                       & 88.2                & {\ul \textbf{79.9}}        & {\ul \textbf{71.6}}        & {\ul 76.3}                 & 80.3                \\ \cline{1-1}
		AGF-t$^B$                & \textbf{60.4}              & 43.8                       & \textbf{55.9}              & \textbf{60.3} & —                          & —                          & —                          & —                   & —                          & —                          & —                          & —                   \\ \hline
		AGF-p$^{BF}$            & 63.4                       & \textbf{48.6}              & 55.5                       & 62.9          & 83.7                       & 76.0                       & 81.3                       & 90.7                & 79.6                       & \textbf{73.8}              & \textbf{77.1}              & 81.2                \\ \cline{1-1}
		AGF$^{BF}$              & 63.2                       & {\ul 48.0}                 & {\ul 54.5}                 & {\ul 62.6}    & {\ul \textbf{83.8}}        & {\ul \textbf{77.0}}        & {\ul \textbf{81.5}}        & {\ul \textbf{91.2}} & 79.1                       & 73.2                       & {\ul 76.2}                 & {\ul \textbf{81.9}} \\ \cline{1-1}
		AGF$_S^{BF}$            & {\ul 63.8}                 & 47.8                       & 54.1                       & 61.6          & 83.1                       & 75.5                       & 80.0                       & 89.8                & {\ul \textbf{80.2}}        & {\ul 73.7}                 & 75.8                       & 81.6                \\ \cline{1-1}
		AGF-t$^{BF}$            & \textbf{64.0}              & 48.0                       & \textbf{55.7}              & \textbf{63.0} & —                          & —                          & —                          & —                   & —                          & —                          & —                          & —                   \\ \hline
	\end{tabular}
\end{table}

\subsection{Ablation Study}

Experimental Results of the variants of AGF are presented in Table~\ref{table:as}. The underlined scores are the better scores between AGF* (i.e., AGF, AGF$^{B}$, AGF$^{BF}$) and AGF$_S$* (i.e., AGF$_S$, AGF$_S^{B}$, AGF$_S^{BF}$). From the results we draw the following conclusions. First, AGF-t* (i.e., AGF-t, AGF-t$^{B}$, AGF-t$^{BF}$) outperform AGF-p* (i.e., AGF-p, AGF-p$^{B}$, AGF-p$^{BF}$) in 11 of 12 results on the ASMOTE task and AGF* surpass AGF-p* on the ATSA task, indicating the effectiveness of SLA. Second, on ASMOTE, AGF* outperform AGF$_S$* in 8 of 12 results, which shows that SLA taking logits as input is a little more effective than SLA taking probabilites as input. Third, AGF* obtain better performances than AGF$_S$* on the ATSA task, while AGF$_S$* obtain better performances than AGF* on TOWE.

\begin{figure}
	\centering
	\includegraphics[width=0.9\textwidth]{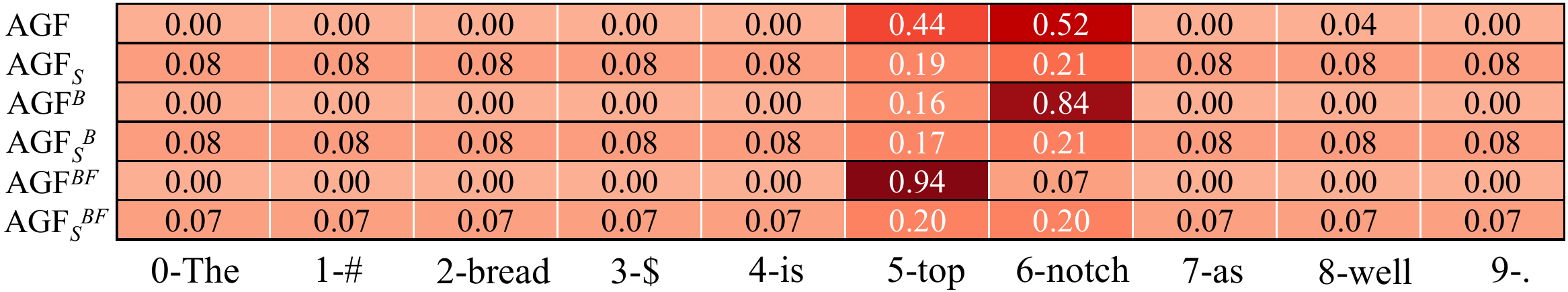}
	\caption{Visualization of attentions.}
	\label{fig:attentions}
\end{figure}

\subsection{Visualization of Attentions}
Fig.~\ref{fig:attentions} shows the attention weights of AGF* and AGF$_S$* on the sentence ``The bread is top notch as well .''. AGF* assign more accurate weights to the opinion words, which can explain why AGF* obtain better performance on the ATSA task. This also can explain why AGF* obtain worse performance on the TOWE task. More accurate weights mean that AGF* are more confident of their prediction on the TOWE task and unfortunately obtain poor generalization ability.

\section{Conclusion}
In this paper, we introduce a new task, Aspect Sentiment Multiple Opinions Triplet Extraction (ASMOTE). One triplet extracted by ASMOTE includes an aspect term, the sentiment toward the aspect term, and all opinion terms associated with the aspect term in the sentence. We build four ASMOTE datasets for the ASMOTE task based on previous ATSA datasets and TOWE datasets. We propose an Aspect-Guided Framework (AGF) with a Sequence Labeling Attention (SLA) for ASMOTE. Moreover, experiments validate the effectiveness of AGF and SLA. These results provide a benchmark performance for ASMOTE.

%
%
%
\bibliographystyle{splncs04}
\bibliography{nlpcc2021.bib,anthology.bib}

\begin{thebibliography}{10}
\providecommand{\url}[1]{\texttt{#1}}
\providecommand{\urlprefix}{URL }
\providecommand{\doi}[1]{https://doi.org/#1}

\bibitem{10.1007/978-3-030-60450-9_52}
Chen, P., Chen, S., Liu, J.: Hierarchical sequence labeling model for aspect
  sentiment triplet extraction. In: Zhu, X., Zhang, M., Hong, Y., He, R. (eds.)
  NLPCC. pp. 654--666. Springer International Publishing, Cham (2020)

\bibitem{chen2020synchronous}
Chen, S., Liu, J., Wang, Y., Zhang, W., Chi, Z.: Synchronous double-channel
  recurrent network for aspect-opinion pair extraction. In: ACL. pp. 6515--6524
  (2020)

\bibitem{chen2020relation}
Chen, Z., Qian, T.: Relation-aware collaborative learning for unified
  aspect-based sentiment analysis. In: ACL. pp. 3685--3694 (2020)

\bibitem{dai2019neural}
Dai, H., Song, Y.: Neural aspect and opinion term extraction with mined rules
  as weak supervision. In: ACL. pp. 5268--5277 (2019)

\bibitem{devlin2019bert}
Devlin, J., Chang, M.W., Lee, K., Toutanova, K.: Bert: Pre-training of deep
  bidirectional transformers for language understanding. In: NAACL-HLT, Volume
  1 (Long and Short Papers). pp. 4171--4186 (2019)

\bibitem{dong2014adaptive}
Dong, L., Wei, F., Tan, C., Tang, D., Zhou, M., Xu, K.: Adaptive recursive
  neural network for target-dependent twitter sentiment classification. In: ACL
  (volume 2: Short papers). pp. 49--54 (2014)

\bibitem{fan2019target}
Fan, Z., Wu, Z., Dai, X., Huang, S., Chen, J.: Target-oriented opinion words
  extraction with target-fused neural sequence labeling. In: NAACL-HLT (2019)

\bibitem{he2019interactive}
He, R., Lee, W.S., Ng, H.T., Dahlmeier, D.: An interactive multi-task learning
  network for end-to-end aspect-based sentiment analysis. In: ACL (2019)

\bibitem{kingma2014adam}
Kingma, D.P., Ba, J.: Adam: A method for stochastic optimization. In: ICLR
  (2015)

\bibitem{li2019unified}
Li, X., Bing, L., Li, P., Lam, W.: A unified model for opinion target
  extraction and target sentiment prediction. In: AAAI. vol.~33, pp. 6714--6721
  (2019)

\bibitem{li2018aspect}
Li, X., Bing, L., Li, P., Lam, W., Yang, Z.: Aspect term extraction with
  history attention and selective transformation. arXiv preprint
  arXiv:1805.00760  (2018)

\bibitem{li2021finegrained}
Li, Y., Wang, F., Zhang, W., hua Zhong, S., Yin, C., He, Y.: A more
  fine-grained aspect-sentiment-opinion triplet extraction task (2021)

\bibitem{liang2019novel}
Liang, Y., Meng, F., Zhang, J., Xu, J., Chen, Y., Zhou, J.: A novel
  aspect-guided deep transition model for aspect based sentiment analysis. In:
  EMNLP-IJCNLP. pp. 5572--5584 (2019)

\bibitem{liu2012sentiment}
Liu, B.: Sentiment analysis and opinion mining. Synthesis lectures on human
  language technologies  \textbf{5}(1),  1--167 (2012)

\bibitem{pang2008opinion}
Pang, B., Lee, L.: Opinion mining and sentiment analysis. Foundations and
  Trends{\textregistered} in Information Retrieval  \textbf{2}(1--2),  1--135
  (2008)

\bibitem{paszke2017automatic}
Paszke, A., Gross, S., Chintala, S., Chanan, G., Yang, E., DeVito, Z., Lin, Z.,
  Desmaison, A., Antiga, L., Lerer, A.: Automatic differentiation in pytorch
  (2017)

\bibitem{Peng2020KnowingWH}
Peng, H., Xu, L., Bing, L., Huang, F., Lu, W., Si, L.: Knowing what, how and
  why: A near complete solution for aspect-based sentiment analysis. In: AAAI
  (2020)

\bibitem{pennington2014glove}
Pennington, J., Socher, R., Manning, C.D.: Glove: Global vectors for word
  representation. In: EMNLP. pp. 1532--1543 (2014)

\bibitem{phan2020modelling}
Phan, M.H., Ogunbona, P.O.: Modelling context and syntactical features for
  aspect-based sentiment analysis. In: ACL. pp. 3211--3220 (2020)

\bibitem{pontiki-etal-2016-semeval}
Pontiki, M., Galanis, D., Papageorgiou, H., Androutsopoulos, I., Manandhar, S.,
  AL-Smadi, M., Al-Ayyoub, M., Zhao, Y., Qin, B., De~Clercq, O., Hoste, V.,
  Apidianaki, M., Tannier, X., Loukachevitch, N., Kotelnikov, E., Bel, N.,
  Jim{\'e}nez-Zafra, S.M., Eryi{\u{g}}it, G.: {S}em{E}val-2016 task 5: Aspect
  based sentiment analysis. In: {S}em{E}val-2016. pp. 19--30. ACL, San Diego,
  California (Jun 2016)

\bibitem{pontiki-etal-2015-semeval}
Pontiki, M., Galanis, D., Papageorgiou, H., Manandhar, S., Androutsopoulos, I.:
  {S}em{E}val-2015 task 12: Aspect based sentiment analysis. In: {S}em{E}val
  2015. pp. 486--495. ACL, Denver, Colorado (Jun 2015)

\bibitem{pontiki-etal-2014-semeval}
Pontiki, M., Galanis, D., Pavlopoulos, J., Papageorgiou, H., Androutsopoulos,
  I., Manandhar, S.: {S}em{E}val-2014 task 4: Aspect based sentiment analysis.
  In: {S}em{E}val 2014. pp. 27--35. ACL, Dublin, Ireland (Aug 2014)

\bibitem{tang2020dependency}
Tang, H., Ji, D., Li, C., Zhou, Q.: Dependency graph enhanced dual-transformer
  structure for aspect-based sentiment classification. In: ACL. pp. 6578--6588
  (2020)

\bibitem{wang2020relational}
Wang, K., Shen, W., Yang, Y., Quan, X., Wang, R.: Relational graph attention
  network for aspect-based sentiment analysis. arXiv preprint arXiv:2004.12362
  (2020)

\bibitem{wang2016recursive}
Wang, W., Pan, S.J., Dahlmeier, D., Xiao, X.: Recursive neural conditional
  random fields for aspect-based sentiment analysis. arXiv preprint
  arXiv:1603.06679  (2016)

\bibitem{wang2017coupled}
Wang, W., Pan, S.J., Dahlmeier, D., Xiao, X.: Coupled multi-layer attentions
  for co-extraction of aspect and opinion terms. In: AAAI (2017)

\bibitem{wei2020don}
Wei, Z., Hong, Y., Zou, B., Cheng, M., Jianmin, Y.: Don’t eclipse your arts
  due to small discrepancies: Boundary repositioning with a pointer network for
  aspect extraction. In: ACL. pp. 3678--3684 (2020)

\bibitem{wu2019enriching}
Wu, S., He, Y.: Enriching pre-trained language model with entity information
  for relation classification. In: CIKM. pp. 2361--2364 (2019)

\bibitem{wu-etal-2020-grid}
Wu, Z., Ying, C., Zhao, F., Fan, Z., Dai, X., Xia, R.: Grid tagging scheme for
  aspect-oriented fine-grained opinion extraction. In: Findings of EMNLP. pp.
  2576--2585. ACL, Online (Nov 2020)

\bibitem{wu2020latent}
Wu, Z., Zhao, F., Dai, X.Y., Huang, S., Chen, J.: Latent opinions transfer
  network for target-oriented opinion words extraction. In: AAAI (2020)

\bibitem{xing2019earlier}
Xing, B., Liao, L., Song, D., Wang, J., Zhang, F., Wang, Z., Huang, H.: Earlier
  attention? aspect-aware lstm for aspect sentiment analysis. In: IJCAI (2019)

\bibitem{xu2018double}
Xu, H., Liu, B., Shu, L., Philip, S.Y.: Double embeddings and cnn-based
  sequence labeling for aspect extraction. In: ACL. pp. 592--598 (2018)

\bibitem{xu-etal-2020-position}
Xu, L., Li, H., Lu, W., Bing, L.: Position-aware tagging for aspect sentiment
  triplet extraction. In: EMNLP. pp. 2339--2349. ACL, Online (Nov 2020)

\bibitem{zhang-etal-2020-multi-task}
Zhang, C., Li, Q., Song, D., Wang, B.: A multi-task learning framework for
  opinion triplet extraction. In: Findings of EMNLP. pp. 819--828. ACL, Online
  (Nov 2020)

\bibitem{zhao2020spanmlt}
Zhao, H., Huang, L., Zhang, R., Lu, Q., et~al.: Spanmlt: A span-based
  multi-task learning framework for pair-wise aspect and opinion terms
  extraction. In: ACL. pp. 3239--3248 (2020)

\bibitem{zhao2020modeling}
Zhao, P., Hou, L., Wu, O.: Modeling sentiment dependencies with graph
  convolutional networks for aspect-level sentiment classification.
  Knowledge-Based Systems  \textbf{193},  105443 (2020)

\end{thebibliography}
\end{document}